\documentclass[runningheads]{llncs}
\usepackage[T1]{fontenc}
\usepackage{amsmath}
\usepackage{booktabs}
\usepackage{enumitem}
\usepackage{graphicx}
\usepackage{multirow}
\usepackage{hyperref}
\usepackage{subfigure}
\usepackage{wrapfig}
\usepackage{balance}
\usepackage{tcolorbox}
\newtcolorbox{examplebox}{
  colback=gray!5,
  colframe=gray!50,
  arc=4pt,
  boxrule=0.4pt,
  left=6pt,
  right=6pt,
  top=4pt,
  bottom=4pt,
  fontupper=\scriptsize\ttfamily  
} 
\begin{document}
\title{Are a Thousand Words Better Than a Single Picture? Beyond Images - A Framework for Multi-Modal Knowledge Graph Dataset Enrichment}
\titlerunning{Beyond Images}
%
\author{
Pengyu Zhang\inst{1}\orcidID{0000-0001-5111-4487} \and
Klim Zaporojets\inst{2}\orcidID{0000-0003-4988-978X} \and
Jie Liu\inst{1}\orcidID{0000-0001-5116-5401} \and
Jia-Hong Huang\inst{1,3}\orcidID{0000-0001-7943-2591} \and
Paul Groth\inst{1}\orcidID{0000-0003-0183-6910}}
\authorrunning{P. Zhang et al.}
\institute{
University of Amsterdam, Amsterdam, The Netherlands \email{p.zhang@uva.nl} \and
Aarhus University, Aarhus, Denmark \and
Amazon AGI, Seattle, USA
}
\maketitle              
\begin{abstract}
Multi-Modal Knowledge Graphs (MMKGs) benefit from visual information, yet large-scale image collection is hard to curate and often excludes ambiguous but relevant visuals (e.g., logos, symbols, abstract scenes). We present \textbf{Beyond Images}, an automatic data-centric enrichment pipeline with optional human auditing. This pipeline operates in three stages: (1) large-scale retrieval of additional entity-related images, (2) conversion of all visual inputs into textual descriptions to ensure that ambiguous images contribute usable semantics rather than noise, and (3) fusion of multi-source descriptions using a large language model (LLM) to generate concise, entity-aligned summaries. These summaries replace or augment the text modality in standard MMKG models without changing their architectures or loss functions. Across three public MMKG datasets and multiple baseline models, we observe consistent gains (up to \textbf{+7\%} Hits@1 overall). Furthermore, on a challenging subset of entities with visually ambiguous logos and symbols, converting images into text yields large improvements (\textbf{+201.35\%} MRR and \textbf{+333.33\%} Hits@1). Additionally, we release a lightweight Text–Image Consistency Check Interface for optional targeted audits, improving description quality and dataset reliability. Our results show that scaling image coverage and converting ambiguous visuals into text is a practical path to stronger MMKG completion. Code, datasets, and supplementary materials are available at \url{https://github.com/pengyu-zhang/Beyond-Images}.
\end{abstract}
\section{Introduction}
\label{sec:Introduction}

\begin{figure}[t]
\centering
\includegraphics[width=\columnwidth]{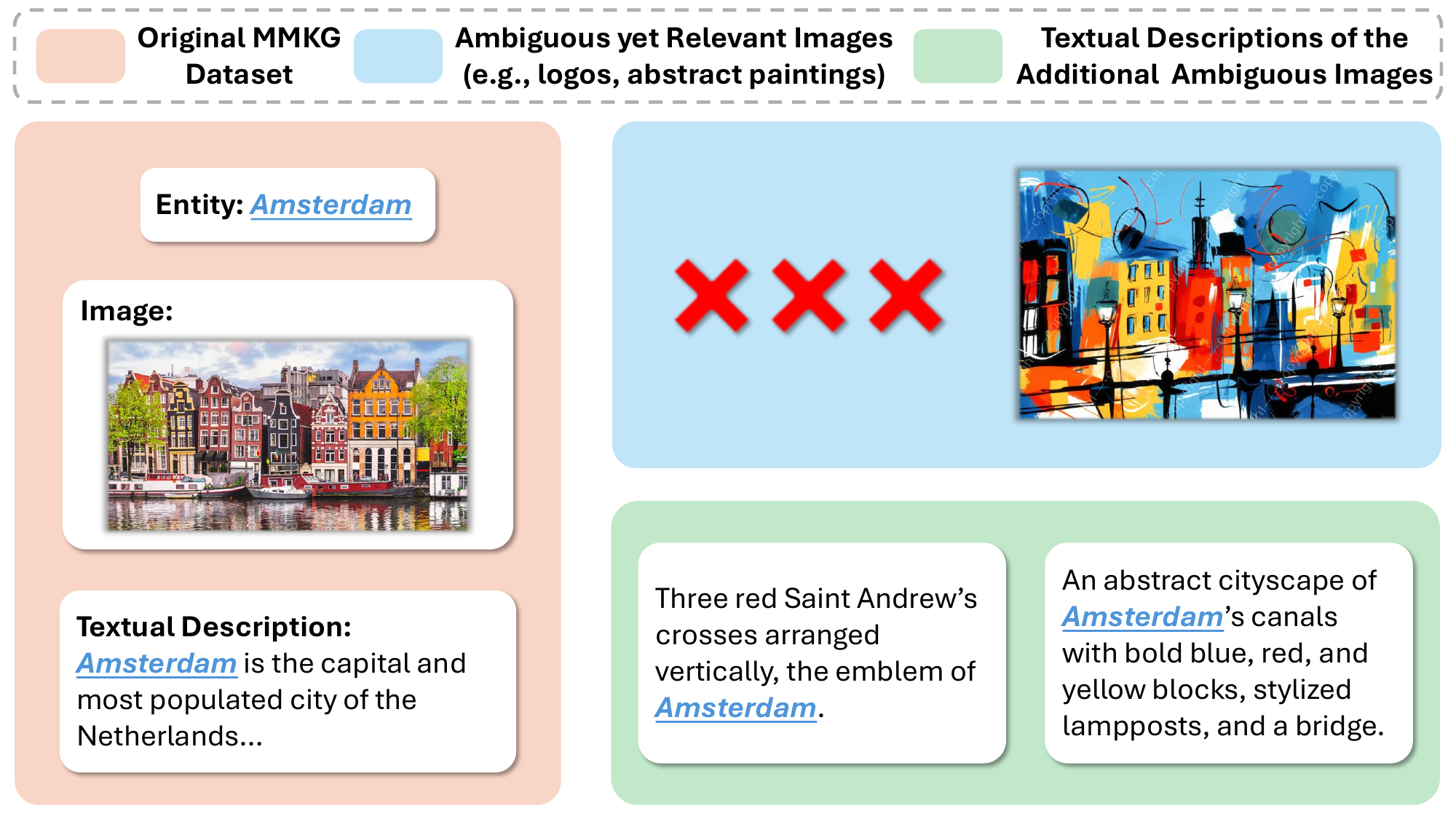}
\caption{\textbf{Original MMKG Dataset}: the entity ``\texttt{Amsterdam}'' with its photo and a textual description. \textbf{Ambiguous yet Relevant Images}: additional visuals such as the three red Saint Andrew's crosses and an abstract cityscape, whose semantics may be unclear when incorporated directly as visual features. \textbf{Textual Descriptions of the Additional Ambiguous Images}: our framework converts these images into concise, entity-aligned text, expanding semantic coverage while mitigating noise from ambiguous visual embeddings.}
\label{figure1}
\end{figure}

In Multi-Modal Knowledge Graphs (MMKGs), integrating text, images, audio, and video with structured triples yields richer signals and improves entity representations \cite{chen2024knowledgegraphsmeetmultimodal,koloski2024automl,10.1145/3587259.3627547}. As illustrated on the left of Figure~\ref{figure1}, the entity ``\texttt{Amsterdam}'' may be associated with textual descriptions of its history and culture, photographs of canals and landmarks, and audio or visual content that reflects its urban atmosphere. Among these modalities, images often convey dense information, and when their content is unambiguous, downstream performance improves \cite{mei2025power}. However, even though the Web enables rapid, large-scale collection of entity-related images, the construction process still suffers from two limitations.

\textbf{Limitation 1: Lack of scalability.} While most existing MMKG datasets automate the acquisition of textual descriptions, image collection often still relies on manual curation \cite{10.1145/3503161.3548388}. Curators or domain experts typically search for each entity and select a small set of images \cite{liu2019mmkg}. This practice helps avoid obviously misleading content but requires manual collection or verification and is difficult to scale to large graphs \cite{li2025fakescope,chen-etal-2024-sac}.

\textbf{Limitation 2: Ambiguous yet relevant images remain unused.} Manual curation typically favors highly representative images for each entity and excludes uncertain or ambiguous images \cite{chen2025ambiguity,10.1007/978-3-031-94575-5_8}. Although this strategy supports quality control, it narrows the dataset's visual coverage and suppresses alternative yet valid information. The Web contains a large number of entity-related images whose visual semantics are inherently ambiguous. When incorporated directly into a dataset, such images often contribute noise rather than useful signal, which lowers data quality and degrades model performance \cite{dayarathna2024deep,KO2025478}. Figure~\ref{figure1} highlights two common forms of image ambiguity: \textit{Sparse-Semantic Images} (e.g., symbolic logos) are visually simple and lack sufficient semantic detail. Although sometimes domain-relevant, they rarely provide distinctive embeddings that benefit MMKG models \cite{su2024loginmea,Wang_Min_Hou_Ma_Zheng_Wang_Jiang_2020}. In contrast, \textit{Rich-Semantic Images} (e.g., abstract paintings) contain visually and semantically complex scenes that current embedding methods struggle to interpret, often leading to significant information loss \cite{wilber2017bam}. Consequently, such images are frequently discarded during dataset construction. While this preserves highly reliable visual signals, it also reduces within-entity diversity. For example, in Figure~\ref{figure1}, for the entity ``\texttt{Amsterdam}'', a dataset may only retain canal photographs but exclude the ``triple red X'' despite its historical and cultural significance to the city.

To address these limitations, we propose an automated framework, \textbf{Beyond Images}, which comprises three stages. First, the \textit{Modality Extension Module} retrieves additional entity-related images beyond the original dataset, enabling large-scale expansion. Next, the \textit{Semantic Generation Module} converts all retrieved images into textual descriptions, ensuring that ambiguous visuals contribute usable semantics rather than noise. Finally, the \textit{LLM-based Semantic Fusion Module} consolidates multi-source descriptions, filters task-irrelevant content, and produces concise, entity-aligned summaries.

Our experiments show that our framework effectively addresses the two limitations defined above. In terms of scalability, we observe consistent improvements across three public MMKG datasets, achieving gains of up to +7\% in Hits@1. To evaluate the impact of ambiguous yet relevant images (e.g., logo-only or symbol-centric visuals), we retain and convert newly retrieved images and measure link-prediction gains from their descriptions. On this challenging subset, performance increases by +201.35\% MRR and +333.33\% Hits@1. Additionally, we assess the quality of the generated image descriptions and show that our fused textual summaries outperform purely generated summaries in terms of overall quality. To support this evaluation, we implement a \textit{Text-Image Consistency Check Interface} that allows for efficient targeted quality assessments by human annotators. Our main contributions are as follows:

\textbf{(i)} We propose \textbf{Beyond Images}, a reusable data-centric enrichment paradigm for MMKGs that converts ambiguous-yet-relevant visuals (e.g., logos or symbolic images) into entity-aligned textual representations and performs entity-level semantic fusion, without modifying downstream model architectures or training objectives.

\textbf{(ii)} We develop a lightweight \textit{Text–Image Consistency Check Interface} to support optional human auditing, enabling low-effort quality inspection of generated entity summaries.

\textbf{(iii)} We release enriched datasets, code, and evaluation protocols, and conduct a systematic study across three datasets on four MMKG models, demonstrating consistent gains, up to +7\% in Hits@1 overall, and +333.33\% in Hits@1 on a challenging subset.

\section{Related Work}

\textbf{Multi-Modal Knowledge Graphs.} Link prediction is a core task in Knowledge Graphs (KGs), aiming to infer missing entities or relations from existing triples. Multi-Modal Knowledge Graphs (MMKGs) extend this setting by combining data from multiple modalities, such as text, images, and numerical features, to enhance representation learning and prediction accuracy. \cite{liu2019mmkg} introduced MMKGs that integrate numerical and visual information, demonstrating improvements in link prediction. Building on this, MCLEA \cite{lin-etal-2022-multi} proposed a contrastive learning framework for multi-modal entity alignment. MCLEA first learns modality-specific representations and then applies contrastive learning to jointly model intra-modal and inter-modal interactions. Subsequently, MMKRL \cite{lu2022mmkrl} incorporated a knowledge reconstruction module to integrate structured and multi-modal data into a unified space. This model also uses adversarial training to enhance robustness and performance. More recently, \cite{lee2024multimodal} introduced the MR-MKG method, which leverages MMKGs to improve reasoning capabilities in large language models. In parallel, \cite{chen2024power} developed the SNAG model, which effectively combines structural, visual, and textual features, yielding improved link prediction performance. \cite{10.1109/TMM.2023.3301279} leverages a multi-modal knowledge graph built from news text and detected visual objects to guide entity-aware image captioning.

Despite these advancements, MMKGs still face notable challenges. Many rely on manual data curation, which limits scalability and can introduce biases. Human experts tend to favor visually straightforward and highly representative images, potentially overlooking others that, although less explicit, could provide valuable additional information about the entity \cite{misra2016seeing}. To address this gap, we propose an automated approach that retrieves and associates entity-related images from external sources without requiring manual annotation. 

\noindent \textbf{Automated Dataset Enrichment.} Automated dataset enrichment is crucial in scaling the construction of MMKGs \cite{10.1007/978-3-031-94575-5_3}. \cite{an-etal-2018-accurate} introduced a method to align textual descriptions with knowledge graph entities, improving the semantic consistency of text embeddings. Building on this idea, \cite{guo2023images} applied vision-language models and LLMs for visual question answering. Further advancements include the ADAGIO framework \cite{xiang-etal-2021-ontoea}, which uses genetic programming to learn efficient augmentation frameworks for knowledge graphs, enhancing data augmentation processes. Similarly, \cite{kuo2022beyond} provides a lightweight automated knowledge graph construction solution by extracting keywords and evaluating relationships using graph Laplacian learning. Lastly, \cite{rezayi-etal-2021-edge} further automated knowledge graph construction from unstructured text by integrating natural language processing techniques for entity extraction and relationship mapping, providing an end-to-end pipeline for converting raw text into structured knowledge. \cite{KO2025478} proposed a multimodal deep-context knowledge extractor that integrates images via hierarchical captions and visual prefixes to enhance named entity recognition (NER) and relation extraction (RE) and build KGs.

However, most existing approaches still involve manual filtering or rely on domain experts to select images, making the process time-consuming and prone to subjective biases. To overcome these limitations, we introduce an automated framework that retrieves, filters, and converts images into textual representations. By transforming visual inputs into entity-aligned text and integrating them through semantic fusion, our approach streamlines dataset construction, reduces reliance on human intervention, and improves both scalability and consistency.

\section{Methodology}

\begin{figure*}[t]
\centering
\includegraphics[width=\columnwidth]{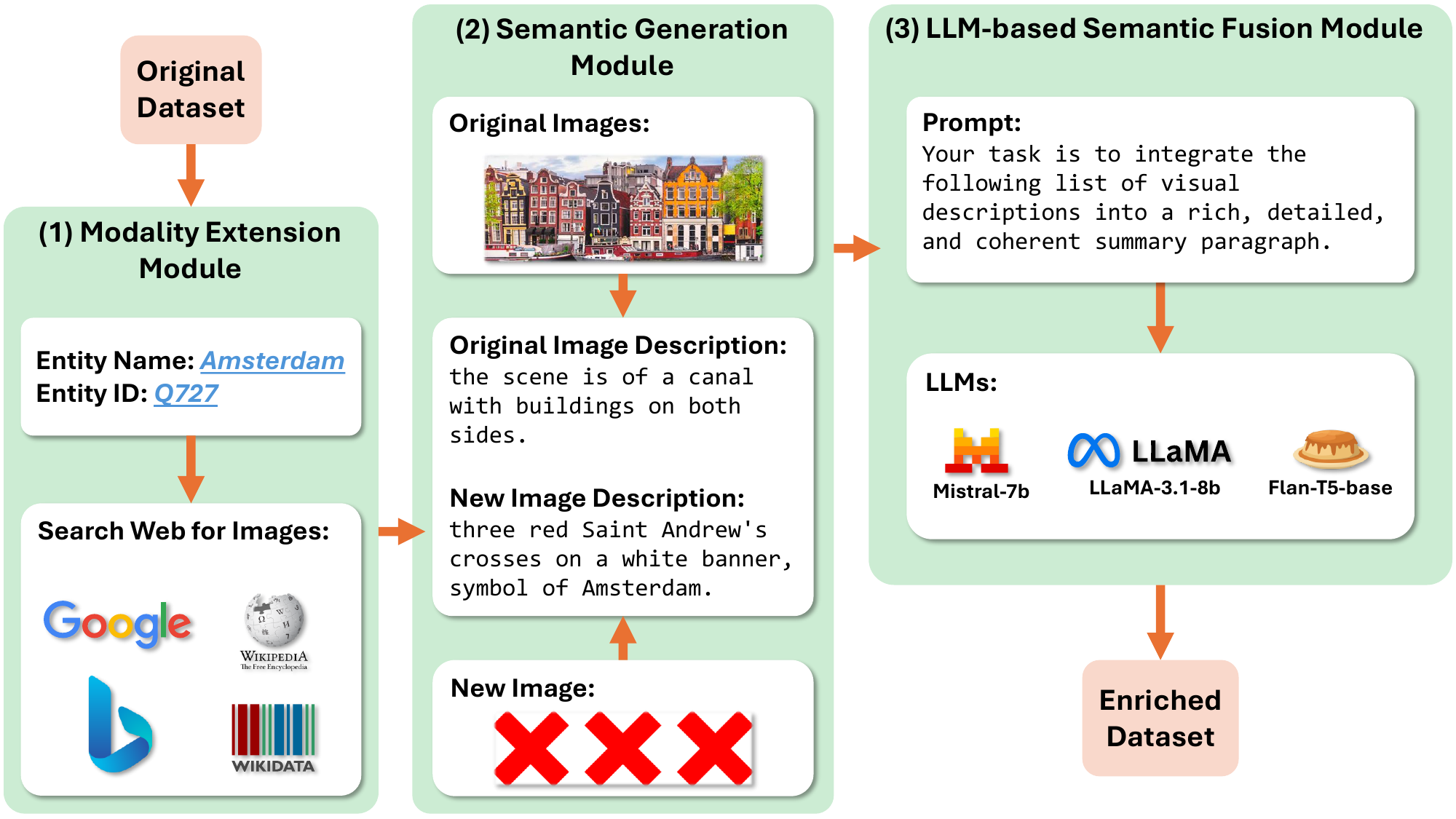}
\caption{Overview of our \textit{Beyond Images} framework. Given an MMKG entity, (1) the \textit{Modality Extension Module} retrieves additional web images from search engines; (2) the \textit{Semantic Generation Module} produces per-image textual descriptions for both original and newly retrieved images; (3) the \textit{LLM-based Semantic Fusion Module} consolidates valid descriptions into a single, rich summary paragraph via an explicit prompt. The summary becomes an enhanced textual view of the entity and is stored in the new enriched dataset for downstream MMKG completion.}
\label{figure2}
\end{figure*}

This section presents our framework, \textbf{Beyond Images}. It consists of three key modules. The \textit{Modality Extension Module} expands the dataset by automatically retrieving additional entity-related images from external sources. The \textit{Semantic Generation Module} then converts both original and the newly retrieved images into textual descriptions to enrich semantic content. Finally, the \textit{LLM-based Semantic Fusion Module} summarizes and filters these descriptions to reduce noise and retain task-relevant information. An overview of the full framework is shown in Figure~\ref{figure2}.

\textbf{Dataset structure (original vs. enriched).} The original MMKG datasets provide: (i) a KG triple set with standard train/validation/test splits, (ii) human-written textual descriptions of the entities (used as the default text modality), and (iii) entity-associated images (raw images or image identifiers/URLs, depending on the dataset). To ensure consistent entity–image alignment, we standardize entity identifiers using Wikidata QIDs and normalize image filenames to the format \texttt{{QID}\_{index}}. Based on the original data, our enriched dataset performs \emph{entity-side attribute enrichment} without modifying the underlying triple structure. Specifically, we add: (i) automatically generated captions for the original images, (ii) automatically generated captions for newly retrieved images when the retrieval module is enabled, (iii) an entity-level fused summary produced by an LLM over all available captions for the entity (original + retrieved), and (iv) lightweight provenance metadata for retrieved images (e.g., source page URL, image URL, and extracted fields such as date/author when available). Downstream MMKG models do not consume the images or metadata directly; instead, they use the resulting textual fields as enhanced entity descriptions. Additional details on data processing and implementation are provided in the supplementary material.\footnote{\url{https://github.com/pengyu-zhang/Beyond-Images/tree/main/supplementary_material}}

\subsection{Modality Extension Module}

To enhance the multi-modal coverage of MMKGs, we apply the \textit{Modality Extension Module} to three public datasets: MKG-W,\footnote{\url{https://github.com/quqxui/MMRNS}} MKG-Y,\footnote{\url{https://github.com/quqxui/MMRNS}} and DB15K.\footnote{\url{https://github.com/mniepert/mmkb}} Each dataset contains structured triples, entity descriptions, and images, covering diverse domains such as film, medicine, and geography. Our method is broadly compatible and applicable to any MMKG with visual content. Using entity names from each dataset, we crawled the corresponding English Wikipedia pages and collected all available images. For each image, we also retrieved metadata, including timestamps. Dataset statistics are summarized in Table~\ref{table1}. \textit{Original Images} refer to those already included in the datasets, while \textit{New Images} are retrieved by our framework. Across all datasets, the number of images per entity increases substantially, improving visual diversity and coverage.

\textbf{Applying to other MMKGs.} To run Beyond Images on a new MMKG dataset, we require only an entity list containing entity IDs and entity names (or canonical labels). The pipeline produces (i) per-image captions and (ii) an entity-level fused summary, which can be stored as an additional text attribute for each entity. Key configurable parameters include the number of retrieved images per entity, basic filtering rules (e.g., image size or type), and the choice of captioning model and fusion LLM. Once generated, the enriched textual fields can be reused across downstream models and tasks without additional LLM calls.

\begin{table}[tbp]
\caption{Statistics of the three public MMKG datasets used in our experiments. \textit{Original Images} are the image statistics provided by the source MMKGs. \textit{New Images} are the additional images retrieved by our framework to extend coverage. ``Avg Img'' denotes the average number of images per entity; ``Entity w/ Img'' is the number of entities with at least one image.}
\label{table1}
\centering
\scriptsize
\begin{tabular}{@{}cccc@{}}
\toprule
\textbf{}              & \textbf{MKG-W} & \textbf{MKG-Y} & \textbf{DB15K} \\ \midrule
\textbf{Entity}        & 15,000         & 15,000         & 12,842         \\
\textbf{Relation}      & 169            & 28             & 279            \\
\textbf{Train}         & 34,196         & 21,310         & 79,222         \\
\textbf{Validation}    & 4,276          & 2,665          & 9,902          \\
\textbf{Test}          & 4,274          & 2,663          & 9,904          \\
\textbf{Text}          & 14,123         & 12,305         & 9,078          \\ \midrule
\multicolumn{4}{c}{\textbf{Original Images}}      \\ \midrule
\textbf{Total Img}     & 27,841         & 42,242         & 603,435        \\
\textbf{Avg Img}       & 3.00           & 3.00           & 53.35          \\
\textbf{Entity w/ Img} & 9,285          & 14,099         & 11,311         \\ \midrule
\multicolumn{4}{c}{\textbf{New Images}}           \\ \midrule
\textbf{Total Img}     & 81,323         & 56,646         & 176,858        \\
\textbf{Avg Img}       & 5.81           & 4.23           & 14.58          \\
\textbf{Entity w/ Img} & 14,002         & 14,388         & 12,130         \\ \bottomrule
\end{tabular}
\end{table}

\subsection{Semantic Generation Module}

To capture semantic information from both original and newly collected multi-modal data, we introduce the \textit{Semantic Generation Module}. This module uses three state-of-the-art image-to-text models: ``\textit{blip2-flan-t5-xxl}''\footnote{\url{https://huggingface.co/Salesforce/blip2-flan-t5-xxl}} \cite{pmlr-v202-li23q}, ``\textit{git-large-coco}''\footnote{\url{https://huggingface.co/microsoft/git-large-coco}} \cite{wang2022gitgenerativeimagetotexttransformer}, and ``\textit{llava-v1.5-7b}''\footnote{\url{https://huggingface.co/liuhaotian/llava-v1.5-7b}} \cite{lin2024moellavamixtureexpertslarge} to generate rich textual descriptions from images. These models are selected for their strong generative capabilities and their ability to produce detailed, textual descriptions. In contrast, we exclude encoder-only models such as ``\textit{CLIP}''\footnote{\url{https://github.com/openai/CLIP}} from our framework, as they lack generative decoder and therefore cannot produce free-form text.

For each image, we generate a descriptive sentence using each of the three models and associate it with the corresponding entity in the dataset. To ensure simplicity and reproducibility, we use a single fixed prompt for all images: ``Describe the scene, objects, colors, and other details in detail.'' We do not perform a prompt-level ablation in this paper; we discuss this as a limitation in Section~\ref{sec:Limitations}.

\subsection{LLM-based Semantic Fusion Module}

While image-to-text models enrich multi-modal inputs with semantic details, they often introduce noise and task-irrelevant content. For instance, vision-language models can generate repetitive tokens (e.g., ``person, person, person...'') when processing certain images, or produce vague and redundant descriptions when the image is only loosely related to the target entity. This issue is particularly noticeable for entities associated with countries where the retrieved images are often maps or diagrams that provide limited value for entity representation learning.

To mitigate this issue, we introduce the \textit{LLM-based Semantic Fusion Module}, which summarizes and aligns the visual descriptions with the semantics of the target entity in the MMKG. Specifically, we employ three LLMs: ``\textit{Flan-T5-base}'',\footnote{\url{https://huggingface.co/google/flan-t5-base}} ``\textit{LLaMA-3.1-8b-instruct}'',\footnote{\url{https://huggingface.co/meta-llama/Llama-3.1-8B-Instruct}} and ``\textit{Mistral-7b-instruct-v0.3}'',\footnote{\url{https://huggingface.co/mistralai/Mistral-7B-Instruct-v0.3}} to summarize and filter the generated descriptions.

For each entity, we collect all image-based textual descriptions and feed them into the LLM using the following prompt: ``Your task is to integrate the following list of visual descriptions for the entity `\{entity\_name\}' into a rich, detailed, and coherent summary paragraph. Capture as many key details as possible, such as objects, colors, actions, and settings. Your final output must be a single paragraph, not a list.'' This process enables the LLM to extract informative features, remove redundancy, and produce a more coherent and entity-specific semantic summary. As a result, it improves the quality and relevance of the multi-modal inputs used for representation learning.

\section{Experiments}

This section presents our experimental setup and provides a comprehensive evaluation of the proposed framework on three widely used public MMKG datasets using four different model variants. Due to space constraints, implementation details are provided in the supplementary material.\footnote{\url{https://github.com/pengyu-zhang/Beyond-Images/tree/main/supplementary_material}} Our evaluation focuses on the following five research questions (RQs):

\textbf{RQ1 (Performance).} Does the semantically enhanced dataset improve MMKG completion performance, particularly on ambiguous yet relevant image subsets? (Section~\ref{sec:Overall Effectiveness (RQ1)})

\textbf{RQ2 (Description Quality).} What is the quality of the generated textual descriptions? (Section~\ref{sec:Description Quality (RQ2)})

\textbf{RQ3 (Ablation Studies).} Can textual descriptions generated from images serve as an effective substitute for image embeddings? (Section~\ref{sec:Modality Contribution (RQ3)})

\textbf{RQ4 (Parameter Sensitivity).} How do different image-to-text and LLM models influence performance, and how sensitive is the framework to these choices? (Section~\ref{sec:Impact of Model Choices (RQ4)})

\textbf{RQ5 (Case Study).} Which types of triples benefit most from image-generated textual descriptions, and how does our framework affect these cases? (Section~\ref{sec:Case Analysis (RQ5)})

\subsection{Evaluated Models}

To ensure a thorough evaluation, we select four recent and widely used Multi-Modal Knowledge Graph (MMKG) models: \textbf{MMRNS}\footnote{\url{https://github.com/quqxui/MMRNS}} \cite{10.1145/3503161.3548388}, \textbf{MyGO}\footnote{\url{https://github.com/zjukg/MyGO}} \cite{zhang2024mygo}, \textbf{NativE}\footnote{\url{https://github.com/zjukg/NATIVE}} \cite{zhang2024native}, and \textbf{AdaMF}\footnote{\url{https://github.com/zjukg/AdaMF-MAT}} \cite{zhang2024unleashing}. These models are chosen because they operate on datasets that include the original images, enabling a fair and consistent comparison. In addition, they are publicly available, widely recognized, and representative of recent progress in MMKG research. Rather than proposing a new model, our goal is to evaluate a \emph{data-side} enrichment intervention. We therefore test the same enriched inputs across \textbf{four} MMKG architectures, and further vary the enrichment components by using \textbf{three} image-to-text captioning models and \textbf{three} fusion LLMs, all under the same training and evaluation protocols.

\subsection{Task and Evaluation Setup}

We evaluate link prediction on MMKGs. For each test triple, we construct two queries:
\textit{tail prediction} $(h,r,?)$ and \textit{head prediction} $(?,r,t)$. Given a query, the model scores all candidate entities $e\!\in\!\mathcal{E}$ and ranks them by the compatibility score $s(h,r,e)$ or $s(e,r,t)$.

\textbf{The use of enriched descriptions.} Our framework operates entirely on the data side. LLMs are used offline to convert images into text and to fuse multiple captions into a single entity-level summary. During training and inference, no LLM prompting is performed. The resulting summary is used as the textual modality input to standard MMKG models (e.g., MMRNS, MyGO, NativE, AdaMF) without changing their architectures or loss functions.

\textbf{Metrics.} We report Mean Reciprocal Rank (MRR) and Hits@$K$ with $K\in\{1,3,10\}$, averaged over head and tail queries. MRR is the mean of $1/\mathrm{rank}$ of the correct entity. Hits@$K$ measures the fraction of queries where the correct entity appears in the top-$K$ results. Higher values indicate better performance.

\subsection{Overall Effectiveness (RQ1)}
\label{sec:Overall Effectiveness (RQ1)}

\textbf{Overall Performance.} We use the ``\textit{blip2-flan-t5-xxl}'' model to generate textual descriptions from images and then apply ``\textit{Mistral-7b-instruct-v0.3}'' to summarize all descriptions associated with each entity. This combination was selected due to its superior overall performance. A detailed analysis of how different image-to-text models and LLMs influence performance is provided in Section~\ref{sec:Modality Contribution (RQ3)}. The key link prediction results are reported in Table~\ref{table2}, which compares the performance of four MMKG models (MMRNS, MyGO, NativE, and AdaMF) across three datasets (MKG-W, MKG-Y, and DB15K) under different input settings.

\begin{table}[t]
\caption{Link prediction results on three MMKG datasets with four models (MMRNS, MyGO, NativE, AdaMF). Rows denote input settings: \textit{Baseline} (original MMKG only), \textit{G(o)} (textual descriptions from original images), \textit{G(n)} (textual descriptions from newly retrieved images), \textit{G(o+n)} (concatenation of \textit{G(o)} and \textit{G(n)}), and \textit{Fusion} (an LLM summary generated from all image descriptions). Columns report MRR and Hits@$K$. Bold numbers indicate the best setting per model and dataset. The last row, \textit{improv.(\%)}, gives the relative gain of \textit{Fusion} over the \textit{Baseline} for each metric.}

\label{table2}
\centering
\scriptsize
\begin{tabular}{@{}ccccccccccccc@{}}
\toprule
\multicolumn{1}{c|}{}                               & \multicolumn{4}{c|}{\textbf{MKG-W}}                                                    & \multicolumn{4}{c|}{\textbf{MKG-Y}}                                                    & \multicolumn{4}{c}{\textbf{DB15K}}                                \\
\multicolumn{1}{c|}{\multirow{-2}{*}{}}             & \textbf{MRR}   & \textbf{H@1}   & \textbf{H@3}   & \multicolumn{1}{c|}{\textbf{H@10}}  & \textbf{MRR}   & \textbf{H@1}   & \textbf{H@3}   & \multicolumn{1}{c|}{\textbf{H@10}}  & \textbf{MRR}   & \textbf{H@1}   & \textbf{H@3}   & \textbf{H@10}  \\ \midrule
\multicolumn{13}{c}{\textbf{MMRNS \cite{10.1145/3503161.3548388}}}                                                                                                                                                                                                                                               \\ \midrule
\multicolumn{1}{c|}{\textbf{Baseline}}              & 35.03          & 28.59          & 37.49          & \multicolumn{1}{c|}{47.47}          & 35.93          & 30.53          & 39.07          & \multicolumn{1}{c|}{45.47}          & 32.68          & 23.01          & 37.86          & 51.01          \\
\multicolumn{1}{c|}{\textbf{G(o)}}                  & 35.73          & 29.65          & 38.37          & \multicolumn{1}{c|}{48.69}          & 36.59          & 31.78          & 40.19          & \multicolumn{1}{c|}{46.43}          & 33.57          & 24.04          & 39.13          & 52.71          \\
\multicolumn{1}{c|}{\textbf{G(n)}}                  & 36.13          & 29.93          & 38.58          & \multicolumn{1}{c|}{49.02}          & 36.93          & 31.96          & 40.33          & \multicolumn{1}{c|}{46.58}          & 33.37          & 23.78          & 39.02          & 52.40          \\
\multicolumn{1}{c|}{\textbf{G(o+n)}}                & 36.26          & 30.08          & 38.70          & \multicolumn{1}{c|}{49.19}          & 37.03          & 32.12          & 40.46          & \multicolumn{1}{c|}{46.70}          & 33.67          & 24.16          & 39.27          & 52.89          \\
\multicolumn{1}{c|}{\textbf{Fusion}}                & \textbf{37.04} & \textbf{30.54} & \textbf{39.05} & \multicolumn{1}{c|}{\textbf{49.96}} & \textbf{37.54} & \textbf{32.75} & \textbf{40.86} & \multicolumn{1}{c|}{\textbf{47.32}} & \textbf{34.47} & \textbf{24.70} & \textbf{39.70} & \textbf{53.47} \\
\multicolumn{1}{c|}{\textit{\textbf{improv.(\%)}}} & \textbf{\texttt{+}5.74}  & \textbf{\texttt{+}6.82}  & \textbf{\texttt{+}4.15}  & \multicolumn{1}{c|}{\textbf{\texttt{+}5.24}}  & \textbf{\texttt{+}4.48}  & \textbf{\texttt{+}7.26}  & \textbf{\texttt{+}4.58}  & \multicolumn{1}{c|}{\textbf{\texttt{+}4.06}}  & \textbf{\texttt{+}5.47}  & \textbf{\texttt{+}7.33}  & \textbf{\texttt{+}4.86}  & \textbf{\texttt{+}4.82}  \\ \midrule
\multicolumn{13}{c}{\textbf{MyGO \cite{zhang2024mygo}}}                                                                                                                                                                                                                                                \\ \midrule
\multicolumn{1}{c|}{\textbf{Baseline}}              & 36.10          & 29.78          & 38.54          & \multicolumn{1}{c|}{47.75}          & 38.51          & 33.39          & 39.03          & \multicolumn{1}{c|}{47.87}          & 37.72          & 30.08          & 41.26          & 52.21          \\
\multicolumn{1}{c|}{\textbf{G(o)}}                  & 37.19          & 30.85          & 39.65          & \multicolumn{1}{c|}{48.75}          & 39.63          & 34.73          & 39.88          & \multicolumn{1}{c|}{48.90}          & 38.84          & 31.53          & 42.37          & 53.74          \\
\multicolumn{1}{c|}{\textbf{G(n)}}                  & 37.28          & 31.26          & 39.74          & \multicolumn{1}{c|}{49.18}          & 39.83          & 35.07          & 40.20          & \multicolumn{1}{c|}{49.22}          & 38.77          & 31.23          & 42.30          & 53.24          \\
\multicolumn{1}{c|}{\textbf{G(o+n)}}                & 37.42          & 31.42          & 39.88          & \multicolumn{1}{c|}{49.35}          & 39.97          & 35.26          & 40.32          & \multicolumn{1}{c|}{49.37}          & 38.97          & 31.69          & 42.49          & 53.92          \\
\multicolumn{1}{c|}{\textbf{Fusion}}                & \textbf{38.05} & \textbf{31.82} & \textbf{40.54} & \multicolumn{1}{c|}{\textbf{49.82}} & \textbf{40.77} & \textbf{35.69} & \textbf{40.78} & \multicolumn{1}{c|}{\textbf{50.01}} & \textbf{39.38} & \textbf{32.24} & \textbf{43.06} & \textbf{54.22} \\
\multicolumn{1}{c|}{\textit{\textbf{improv.(\%)}}} & \textbf{\texttt{+}5.41}  & \textbf{\texttt{+}6.86}  & \textbf{\texttt{+}5.19}  & \multicolumn{1}{c|}{\textbf{\texttt{+}4.34}}  & \textbf{\texttt{+}5.87}  & \textbf{\texttt{+}6.89}  & \textbf{\texttt{+}4.47}  & \multicolumn{1}{c|}{\textbf{\texttt{+}4.47}}  & \textbf{\texttt{+}4.40}  & \textbf{\texttt{+}7.19}  & \textbf{\texttt{+}4.36}  & \textbf{\texttt{+}3.85}  \\ \midrule
\multicolumn{13}{c}{\textbf{NativE \cite{zhang2024native}}}                                                                                                                                                                                                                                              \\ \midrule
\multicolumn{1}{c|}{\textbf{Baseline}}              & 36.58          & 29.56          & 39.65          & \multicolumn{1}{c|}{48.94}          & 39.04          & 34.79          & 40.89          & \multicolumn{1}{c|}{46.18}          & 37.16          & 28.01          & 41.36          & 54.13          \\
\multicolumn{1}{c|}{\textbf{G(o)}}                  & 37.37          & 30.56          & 40.44          & \multicolumn{1}{c|}{49.93}          & 39.63          & 35.95          & 41.93          & \multicolumn{1}{c|}{47.03}          & 38.68          & 28.83          & 42.48          & 55.11          \\
\multicolumn{1}{c|}{\textbf{G(n)}}                  & 37.57          & 30.68          & 40.85          & \multicolumn{1}{c|}{50.27}          & 39.75          & 36.12          & 42.11          & \multicolumn{1}{c|}{47.43}          & 38.30          & 28.77          & 42.35          & 55.03          \\
\multicolumn{1}{c|}{\textbf{G(o+n)}}                & 37.69          & 30.80          & 40.97          & \multicolumn{1}{c|}{50.41}          & 39.83          & 36.27          & 42.25          & \multicolumn{1}{c|}{47.56}          & 38.84          & 28.92          & 42.61          & 55.22          \\
\multicolumn{1}{c|}{\textbf{Fusion}}                & \textbf{38.04} & \textbf{31.43} & \textbf{41.42} & \multicolumn{1}{c|}{\textbf{51.04}} & \textbf{40.38} & \textbf{36.92} & \textbf{42.72} & \multicolumn{1}{c|}{\textbf{48.30}} & \textbf{39.55} & \textbf{29.37} & \textbf{43.12} & \textbf{55.62} \\
\multicolumn{1}{c|}{\textit{\textbf{improv.(\%)}}} & \textbf{\texttt{+}3.98}  & \textbf{\texttt{+}6.33}  & \textbf{\texttt{+}4.45}  & \multicolumn{1}{c|}{\textbf{\texttt{+}4.30}}  & \textbf{\texttt{+}3.43}  & \textbf{\texttt{+}6.13}  & \textbf{\texttt{+}4.47}  & \multicolumn{1}{c|}{\textbf{\texttt{+}4.59}}  & \textbf{\texttt{+}6.42}  & \textbf{\texttt{+}4.84}  & \textbf{\texttt{+}4.27}  & \textbf{\texttt{+}2.76}  \\ \midrule
\multicolumn{13}{c}{\textbf{AdaMF \cite{zhang2024unleashing}}}                                                                                                                                                                                                                                               \\ \midrule
\multicolumn{1}{c|}{\textbf{Baseline}}              & 35.85          & 29.04          & 39.01          & \multicolumn{1}{c|}{48.42}          & 38.57          & 34.34          & 40.59          & \multicolumn{1}{c|}{45.76}          & 35.14          & 25.30          & 41.11          & 52.92          \\
\multicolumn{1}{c|}{\textbf{G(o)}}                  & 36.92          & 30.16          & 39.78          & \multicolumn{1}{c|}{49.34}          & 39.79          & 35.37          & 41.45          & \multicolumn{1}{c|}{46.41}          & 36.20          & 26.24          & 42.29          & 54.35          \\
\multicolumn{1}{c|}{\textbf{G(n)}}                  & 37.20          & 30.35          & 39.77          & \multicolumn{1}{c|}{49.73}          & 40.05          & 35.86          & 41.89          & \multicolumn{1}{c|}{46.78}          & 35.85          & 26.08          & 42.13          & 54.24          \\
\multicolumn{1}{c|}{\textbf{G(o+n)}}                & 37.36          & 30.50          & 39.85          & \multicolumn{1}{c|}{49.88}          & 40.21          & 36.04          & 42.02          & \multicolumn{1}{c|}{46.88}          & 36.32          & 26.34          & 42.43          & 54.51          \\
\multicolumn{1}{c|}{\textbf{Fusion}}                & \textbf{38.04} & \textbf{31.29} & \textbf{40.39} & \multicolumn{1}{c|}{\textbf{50.46}} & \textbf{40.54} & \textbf{36.68} & \textbf{42.36} & \multicolumn{1}{c|}{\textbf{47.48}} & \textbf{36.74} & \textbf{26.98} & \textbf{42.89} & \textbf{54.84} \\
\multicolumn{1}{c|}{\textit{\textbf{improv.(\%)}}} & \textbf{\texttt{+}6.11}  & \textbf{\texttt{+}7.76}  & \textbf{\texttt{+}3.54}  & \multicolumn{1}{c|}{\textbf{\texttt{+}4.21}}  & \textbf{\texttt{+}5.10}  & \textbf{\texttt{+}6.82}  & \textbf{\texttt{+}4.37}  & \multicolumn{1}{c|}{\textbf{\texttt{+}3.77}}  & \textbf{\texttt{+}4.56}  & \textbf{\texttt{+}6.63}  & \textbf{\texttt{+}4.33}  & \textbf{\texttt{+}3.63}  \\ \bottomrule
\end{tabular}
\end{table}

In Table~\ref{table2}, the first row for each model reports results on the original dataset, reproduced from the corresponding papers. These results use only the default textual descriptions and image embeddings provided in the original datasets. 
The remaining rows evaluate four configurations, each adding a single additional input to the original dataset: ``\textit{G(o)}'', ``\textit{G(n)}'', ``\textit{G(o+n)}'', or ``\textit{Fusion}''. ``\textit{G(o)}'' uses textual descriptions generated from the original images. ``\textit{G(n)}'' uses textual descriptions generated from newly downloaded images. ``\textit{G(o+n)}'' combines both sources by concatenating their descriptions. ``\textit{Fusion}'' uses an LLM to summarize all descriptions from both original and new images into a single paragraph, which is then used as input. ``\textit{H@$K$}'' refers to Hits at $K$, and ``\textit{improv.(\%)}'' indicates the relative performance gain, computed as: \( \text{Boost} = \frac{\text{Fusion Result} - \text{Baseline Result}}{\text{Baseline Result}} \).

The results in Table~\ref{table2} show that using the enriched datasets consistently improves performance across all metrics (MRR, Hits@1, Hits@3, and Hits@10) for every model. For example, MyGO achieves 5.41\% and 6.86\% improvements in MRR and Hits@1 respectively on the MKG-W dataset. Similar improvements are observed on MKG-Y and DB15K, indicating that the proposed approach generalizes across different models and datasets. These results confirm the benefits of incorporating image-based textual descriptions into MMKG tasks.

Among all configurations, ``\textit{Fusion}'' consistently achieves the best performance. Compared with the simple concatenation strategy in ``\textit{G(o+n)}'', the LLM-generated summary provides additional gains, suggesting that naive aggregation fails to filter out noise and redundancy. In contrast, summarizing the image descriptions using an LLM produces more coherent and semantically aligned content with respect to the entity, leading to improved downstream performance.

It is worth noting that while ``\textit{Fusion}'' performs best on all datasets, the comparison between ``\textit{G(o)}'' and ``\textit{G(n)}'' shows a dataset-specific pattern. On MKG-W and MKG-Y, ``\textit{G(n)}'' outperforms ``\textit{G(o)}'', whereas on DB15K the opposite trend is observed. We attribute this to the fact that DB15K contains more original images than newly downloaded ones (see Table~\ref{table1}). As a result, the model benefits from the larger number of descriptions in ``\textit{G(o)}'', leading to stronger entity representations and higher performance.

\begin{table}[t]
\centering
\scriptsize
\caption{Performance on a subset of entities whose image sets \textit{include logos or symbols}. \textit{Baseline} uses the original MMKG without converting these images. \textit{Fusion} applies our framework to convert the logo-like images into textual descriptions. Results show that our framework recovers usable semantics, yielding large gains in MRR and Hits@$K$.}
\label{table3}
\begin{tabular}{@{}ccccc@{}}
\toprule
                              & \textbf{MRR}    & \textbf{H@1}    & \textbf{H@3}    & \textbf{H@10}   \\ \midrule
\textbf{Baseline}             & 13.89          & 7.50          & 15.00          & 27.50          \\
\textbf{Fusion}        & \textbf{41.87} & \textbf{32.50} & \textbf{47.50} & \textbf{57.50} \\
\textit{\textbf{improv.(\%)}} & \textbf{\texttt{+}201.35} & \textbf{\texttt{+}333.33} & \textbf{\texttt{+}216.67} & \textbf{\texttt{+}109.09} \\ \bottomrule
\end{tabular}
\end{table}

\textbf{Ambiguous yet Relevant Image Subsets.} We manually sampled 20 entities whose images consist of logos or other abstract marks and formed a small evaluation subset as shown in Table~\ref{table3}. We compared two inputs: the ``\textit{Baseline}'', which uses only the original MMKG, and ``\textit{Fusion}'', which converts these logo images into textual descriptions and fuses them with the existing modalities. Although logos provide limited visual information, converting them into text supplies the missing semantics. This yields large gains for link prediction, improving MRR and Hits@$K$ by substantial margins (e.g., +201.35\% MRR, +333.33\% Hits@1).

\textbf{Training Efficiency.} We analyze the training efficiency of the proposed approach. As shown in the supplementary material,\footnote{\url{https://github.com/pengyu-zhang/Beyond-Images/tree/main/supplementary_material}} once generated, the enriched data can be reused across models and tasks without extra cost. Training time increases by only 7-30 minutes, while performance improves, demonstrating a favorable cost-benefit trade-off.

\subsection{Description Quality (RQ2)}
\label{sec:Description Quality (RQ2)}

Our framework performs enrichment automatically in an end-to-end pipeline (retrieval $\rightarrow$ captioning $\rightarrow$ fusion). The following step is optional and serves only to audit the accuracy of the generated summaries. Inspired by the CoT Curation Toolkit,\footnote{\url{https://github.com/caocongfeng/CoT_curation_toolkit}} we release a lightweight browser interface for human verification (demo and code\footnote{\url{https://github.com/pengyu-zhang/Beyond-Images/tree/main/video_demo}}). The interface shows, side by side, the LLM summary and the full image set for a given entity. Annotators select a verdict (Match, Mismatch, or Uncertain), add a brief rationale, and may hide irrelevant images.

From each dataset, we draw a random sample of 100 cases for manual auditing. A case is correct when the summary captures the main visual semantics of the image set; we do not require the summary to identify or verify the entity itself. Under this criterion, we observe no clear mismatches and \textit{two cases} where the description is inaccurate or incomplete. These results indicate that the generated summaries reliably reflect the visual content of the images across datasets.

\subsection{Modality Contribution (RQ3)}
\label{sec:Modality Contribution (RQ3)}

\begin{figure*}[t]
\centering
\subfigure[MMRNS]{\includegraphics[width=0.43\columnwidth]{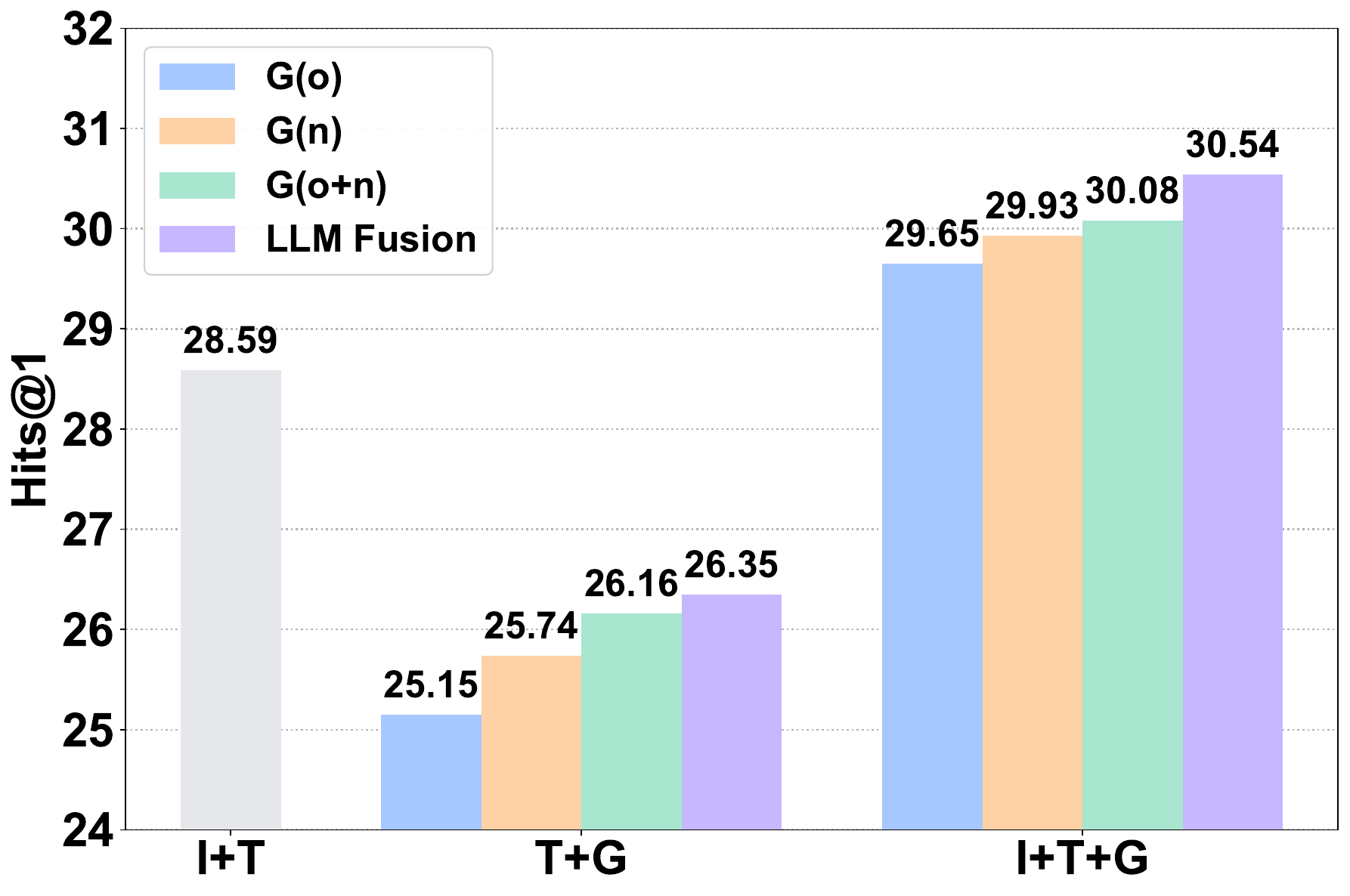}}
\subfigure[MyGO]{\includegraphics[width=0.43\columnwidth]{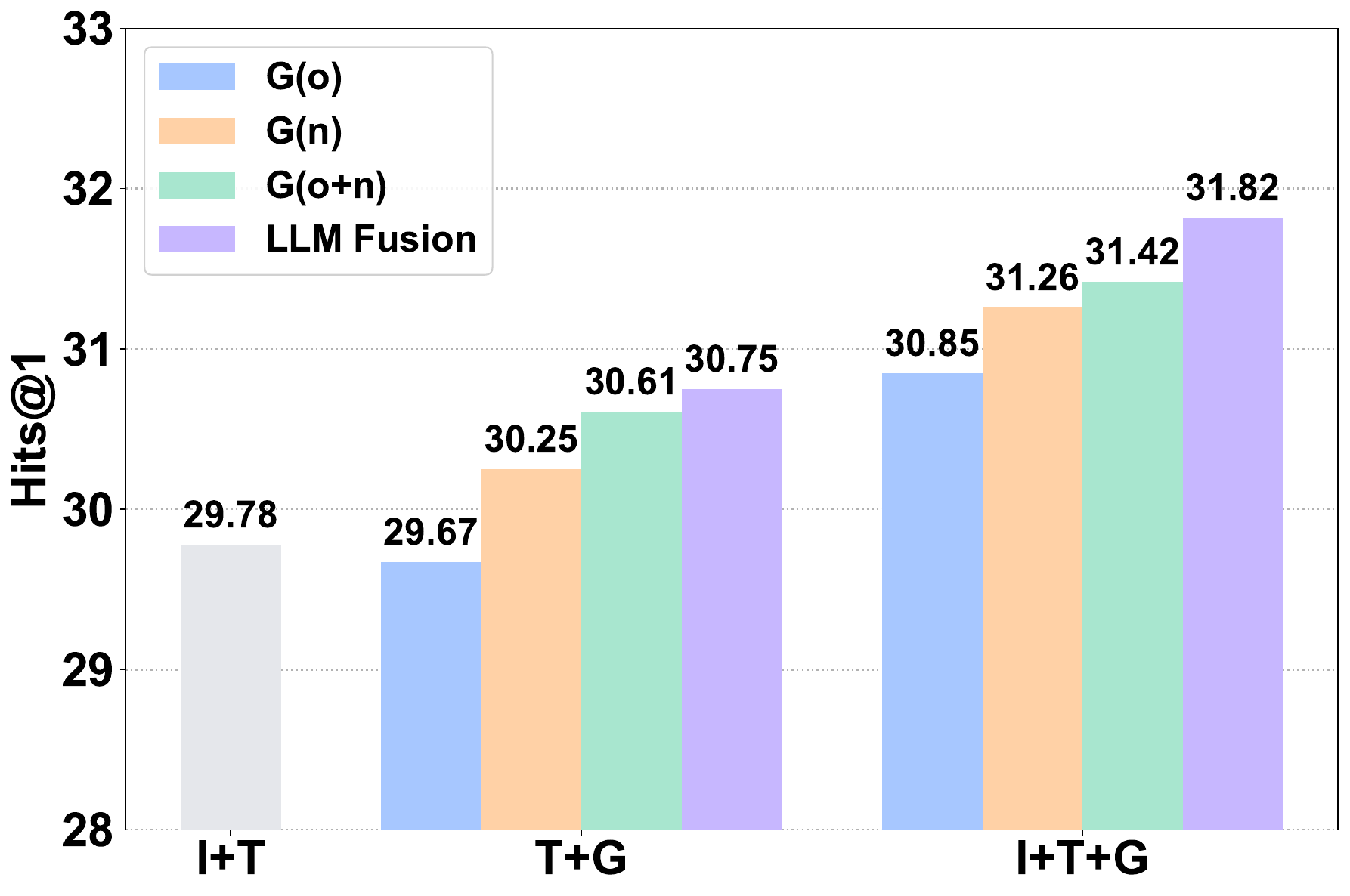}}\\
\subfigure[NativE]{\includegraphics[width=0.43\columnwidth]{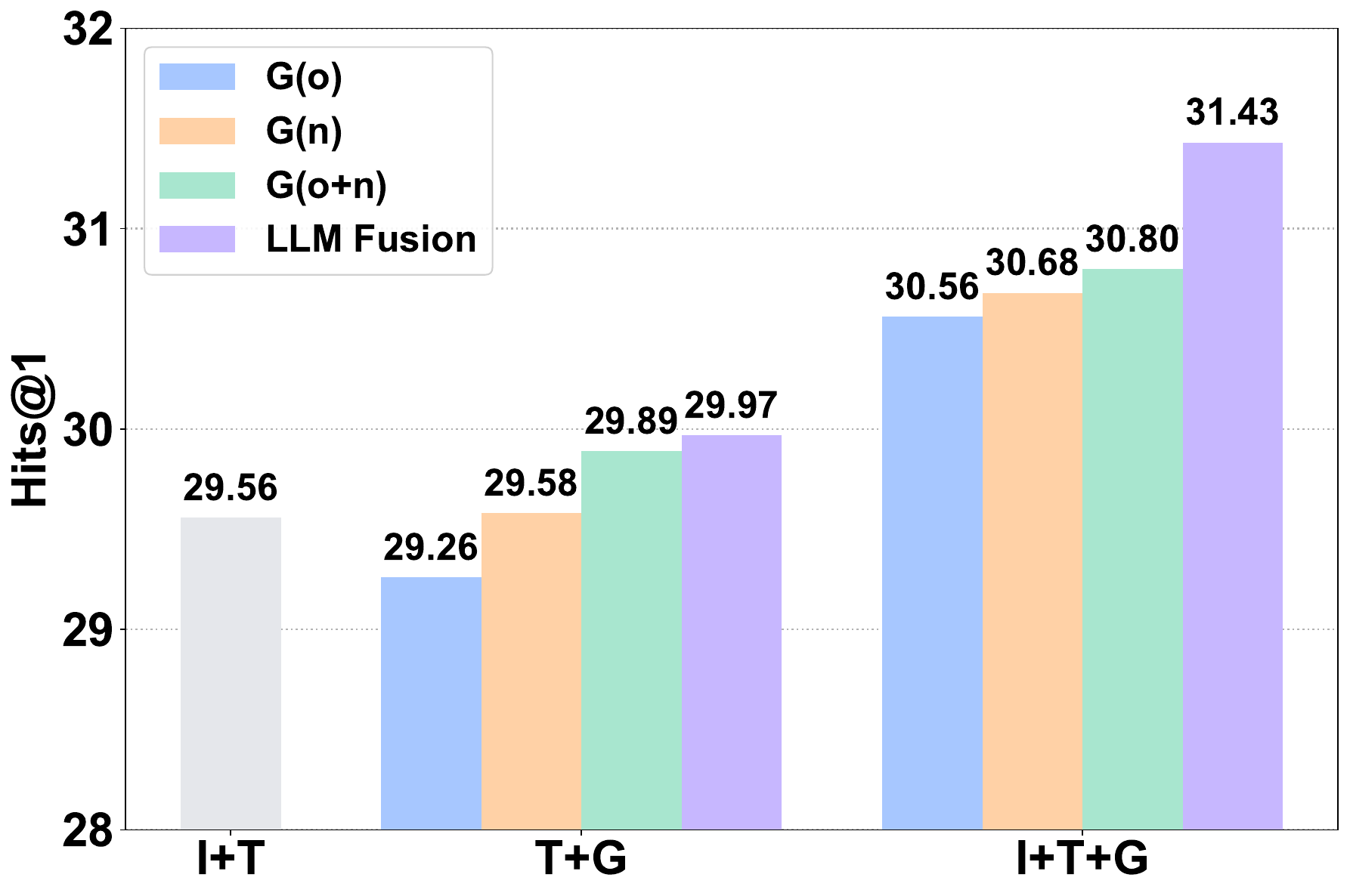}}
\subfigure[AdaMF]{\includegraphics[width=0.43\columnwidth]{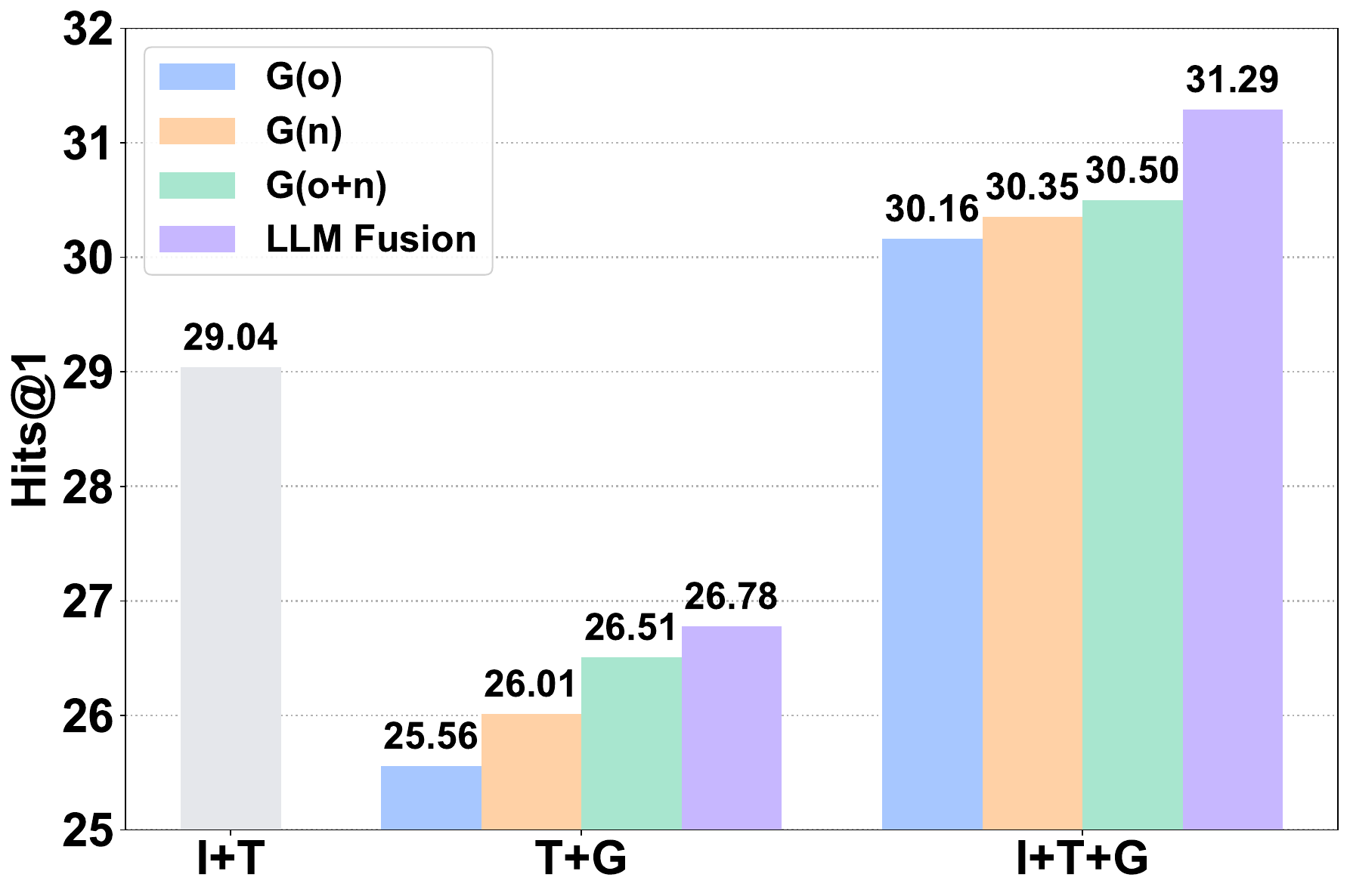}}
\caption{Hits@1 comparison on the MKG-W dataset across four models under three modality settings: \textit{I+T} (image embeddings + textual descriptions), \textit{T+G} (textual descriptions + image-generated descriptions), and \textit{I+T+G} (all). Legend: \textit{G(o)} denotes descriptions from original images, \textit{G(n)} from newly retrieved images, \textit{G(o+n)} their concatenation, and \textit{LLM Fusion} an LLM summary over all descriptions. Bars report Hits@1 (higher is better).}
\label{figure3}
\end{figure*}

Figure~\ref{figure3} reports Hits@1 on the MKG-W dataset for four models (MMRNS, MyGO, NativE, AdaMF) under three different modality combinations (see further). Bars represent Hits@1 (higher is better). The x-axis denotes the modality combinations: \textit{I+T} uses original image embeddings and text; \textit{T+G} uses text plus image-generated descriptions; \textit{I+T+G} uses all three. The legend indicates the source of the image-generated descriptions used in each bar: \textit{G(o)} from original images, \textit{G(n)} from newly retrieved images, \textit{G(o+n)} as their concatenation, and \textit{LLM Fusion} as an LLM summary over all descriptions.

As shown in the figure, using only two modalities results in limited model performance. Performance improves when all three modalities are combined, with the best results observed in the \textit{I+T+G} with \textit{LLM Fusion}. This demonstrates the benefit of integrating complementary information across modalities, which helps close the semantic gap and improves prediction accuracy.

We observe different levels of reliance on visual embeddings versus image generated text across models. For MyGO and NativE, replacing image embeddings with generated descriptions (\textit{T+G} vs. \textit{I+T}) yields comparable or higher Hits@1. In contrast, MMRNS and AdaMF show performance degradation under \textit{T+G} relative to \textit{I+T}, indicating a stronger dependence on visual features. A plausible explanation is architectural: models like MMRNS and AdaMF use global image embeddings that preserve high-level semantics, while models like MyGO and NativE tokenize images into local patches, which can dilute entity-level semantics, making textual descriptions a competitive substitute.

Many MMKG images are \emph{ambiguous yet relevant}, which provides a complementary, data-side explanation for the model-dependent behaviors observed above. For \emph{sparse-semantic} visuals such as logos and symbols, raw visual embeddings can be weak or non-discriminative, making it difficult for a completion model to extract relation-relevant cues. For \emph{rich-semantic} but abstract or stylistic images, fixed visual encoders may miss key contextual details, leading to information loss. In both cases, converting images into text helps make the latent semantics more explicit and easier to align with KG relations, particularly when multiple captions are fused at the entity level to reduce redundancy and noise. Motivated by this intuition, we next conduct a controlled modality ablation to better isolate how much each modality contributes under a fixed architecture.

\begin{table}[t]
\caption{Modality ablation on the MKG-W dataset. We compare \textit{Image Only}, \textit{Text Only}, and \textit{Image + Text}. Text alone outperforms image alone across all metrics, while fusing both modalities yields the best MRR and Hits@$K$.}

\label{table4}
\centering
\scriptsize
\begin{tabular}{@{}ccccc@{}}
\toprule
                & \textbf{MRR}   & \textbf{H@1}   & \textbf{H@3}   & \textbf{H@10}  \\ \midrule
\textbf{Image Only} & 31.70          & 24.47          & 32.34          & 41.68          \\
\textbf{Text Only} & 35.93          & 28.41          & 36.75          & 45.09          \\
\textbf{Image + Text}  & \textbf{36.10} & \textbf{29.78} & \textbf{38.54} & \textbf{47.75} \\ \bottomrule
\end{tabular}
\end{table}

To isolate the contribution of each modality, and inspired by prior work \cite{zhao-etal-2022-mose,li2023imf}, we evaluate the MyGO model on the MKG-W dataset under three settings (as shown in Table~\ref{table4}): \textit{Image Only}, \textit{Text Only} (the original MMKG entity textual descriptions, \textit{not} our image-derived captions), and \textit{Image + Text}. \textit{Image Only} attains the lowest scores (e.g., MRR 31.70; H@1 24.47), indicating that images alone provide limited relational signal for link prediction. \textit{Text Only} improves performance across all metrics (MRR 35.93; H@1 28.41), suggesting that textual descriptions encode stronger cues about entities and relations. \textit{Image + Text} achieves the best performance (MRR 36.10; H@1 29.78), confirming that the two modalities are complementary and that fusing them yields consistent gains. Although the improvements over \textit{Text Only} are moderate, they highlight the incremental yet useful contribution of visual evidence.

\subsection{Impact of Model Choices (RQ4)}
\label{sec:Impact of Model Choices (RQ4)}

In this section, we analyze how the choice of image-to-text models and LLMs affects the overall performance of our framework. We first evaluate the impact of different pre-trained image-to-text models by applying them to the MyGO baseline on the MKG-W dataset. As shown in the upper block of Table~\ref{table5}, all models improve performance compared to the original baseline, with ``\textit{blip2-flan-t5-xxl}'' achieving the highest gains across all metrics.

Next, we study the effect of different LLMs used to summarize the generated image descriptions. Using ``\textit{blip2-flan-t5-xxl}'' as the image-to-text backbone, we compare three LLMs: ``\textit{Flan-T5-base}'', ``\textit{LLaMA-3.1-8b-instruct}'', and ``\textit{Mistral-7b-instruct-v0.3}''. As shown in the lower block of Table~\ref{table5}, all three models outperform the baseline, with ``\textit{Mistral-7b-instruct-v0.3}'' achieving the best results. These findings demonstrate the effectiveness of LLM-based summarization in refining image-derived descriptions and enhancing semantic alignment.

\begin{table}[t]
\caption{Performance comparison of different pre-trained image-to-text generation models on the MKG-W dataset using the MyGO model. ``\textbf{\textit{H@$K$}}'' stands for ``Hits at $K$.''}
\label{table5}
\centering
\scriptsize
\begin{tabular}{@{}ccccc@{}}
\toprule
\textbf{}                         & \textbf{MRR}   & \textbf{H@1}   & \textbf{H@3}   & \textbf{H@10}  \\ \midrule
\multicolumn{5}{c}{\textbf{Image-to-Text Models (no LLM Fusion)}}             \\ \midrule
\textbf{Baseline}                 & 36.10          & 29.78          & 38.54          & 47.75          \\
\textbf{Git-large-coco}           & 36.59          & 30.61          & 38.57          & 48.02          \\
\textbf{Llava-v1.5-7b}            & 36.21          & 30.15          & 38.40          & 47.83          \\
\textbf{Blip2-flan-t5-xxl}        & \textbf{37.42} & \textbf{31.42} & \textbf{39.88} & \textbf{49.35} \\ \midrule
\multicolumn{5}{c}{\textbf{LLM Fusion (using Blip2 for image-to-text)}}       \\ \midrule
\textbf{Baseline}                 & 36.10          & 29.78          & 38.54          & 47.75          \\
\textbf{Flan-T5-base}             & 37.58          & 31.48          & 39.98          & 49.42          \\
\textbf{LLaMA-3.1-8b-instruct}    & 37.87          & 31.56          & 40.02          & 49.61          \\
\textbf{Mistral-7b-instruct-v0.3} & \textbf{38.05} & \textbf{31.82} & \textbf{40.54} & \textbf{49.82} \\ \bottomrule
\end{tabular}
\end{table}

To illustrate these differences, we present a representative example: the album cover of ``\texttt{Hot Sauce Committee Part Two}'' by the ``\texttt{Beastie Boys}''. The ``\textit{blip2-flan-t5-xxl}'' model generates brief yet informative descriptions such as ``The cover of beastboys hot sauce committee part two'' and ``The logo for beastie boys is shown in black and white.'' These outputs capture both the type of image (album cover) and key named entities (artist and album title), which helps the model align visual and textual information more effectively. This alignment improves entity disambiguation and link prediction.

\begin{examplebox}
blip2-flan-t5-xxl: ``The cover of Beastie Boys Hot Sauce Committee Part Two''.
\end{examplebox}

The ``\textit{git-large-coco}'' model only outputs a generic phrase, failing to identify any specific entities. While it correctly recognizes the image type, it lacks the detail needed for fine-grained semantic alignment.

\begin{examplebox}
git-large-coco: ``cover art for the album''.
\end{examplebox}

In comparison, ``\textit{llava-v1.5-7b}'' generates a long and visually rich description, focusing on colors, shapes, and layout.

\begin{examplebox}
llava-v1.5-7b: ``The image is a colorful collage featuring a variety of shapes, colors, and patterns. The collage is composed of multiple squares, each containing different colors and designs. The squares are arranged in a visually appealing manner, creating a vibrant and dynamic composition. The colors in the collage are diverse, with some squares featuring red, green, blue, yellow, and orange hues. The patterns and shapes within the squares vary, adding depth and interest to the overall design. The collage appears to be a creative and artistic representation of the artist's vision, showcasing their talent and skill in combining different elements to create a cohesive and visually engaging piece''.
\end{examplebox}

The different outputs of the three models are likely due to differences in architecture, language modeling, and training objectives. The superior performance of ``\textit{blip2-flan-t5-xxl}'' suggests that generating straightforward and semantically aligned descriptions is beneficial for representation learning and downstream tasks.

\subsection{Case Analysis (RQ5)}
\label{sec:Case Analysis (RQ5)}

To demonstrate how image-generated descriptions enhance model performance, we compare predictions made with the original dataset inputs against those made with our enhanced inputs that additionally include image-generated textual descriptions. The largest gain is observed for the triple \textit{(Hot Sauce Committee Part Two, performer, Beastie Boys)}. Here, the head entity is \textit{Hot Sauce Committee Part Two}, the relation is \textit{performer}, and the tail entity is \textit{Beastie Boys}. With the relation \textit{performer} and the textual description of the tail entity \textit{Beastie Boys}, the rank of the correct head entity improves from 13,680 to 1,330. Likewise, when given the head entity \textit{Hot Sauce Committee Part Two}, its textual description, and the relation \textit{performer}, the rank of the correct tail entity improves from 11,435 to 4,628. This case illustrates that adding image-derived text can strengthen entity alignment by making relation-relevant cues more explicit.

In the original dataset, the associated visuals for this triple are sparse-semantic (e.g., logo-like or symbolic content). Visual embeddings therefore tend to capture abstract shapes and patterns with limited entity-discriminative information, making it difficult for the model to learn meaningful connections from images alone. In contrast, the generated textual descriptions translate such ambiguous visuals into explicit cues (e.g., names or identity hints), which can be more directly aligned with KG relations and thus improve ranking. Due to copyright and reuse restrictions, we do not reproduce the corresponding images in the paper. The entities can be inspected via their public knowledge base pages (e.g., Wikidata\footnote{\url{https://en.wikipedia.org/wiki/File:Hot_Sauce_Committee_Part_Two.png}} \footnote{\url{https://commons.wikimedia.org/wiki/File:Beastie_Boys_logo_(1985-1986).png}}).

\section{Conclusion and Future Work}

We introduced \textbf{Beyond Images}, an automatic enrichment pipeline for MMKGs that scales image retrieval, converts visuals into task-aligned text, and fuses multi-source descriptions into concise, entity-aligned summaries. This \textit{data-side} approach is model-agnostic and can be integrated with standard MMKG models without modifying their architectures or loss functions. Experiments on three public datasets show consistent improvements (up to \textbf{+7\%} Hits@1 overall). On a challenging subset with logo or symbol images, converting images into text substantially boosts performance (\textbf{+201.35\%} MRR; \textbf{+333.33\%} Hits@1), indicating that images with weak visual signals can still convey strong semantics when rendered as text. We additionally provide a lightweight \textit{Text-Image Consistency Check Interface} for optional, low-effort human auditing to further enhance dataset reliability.

For future work, we plan to (i) integrate temporal signals (e.g., timestamps and versioned descriptions) to study entity evolution \cite{NEURIPS2023_066b98e6}, (ii) explore stronger and multilingual captioning and fusion LLMs to extend beyond English-centric corpora, (iii) incorporate active sampling for more efficient human auditing, and (iv) extend the framework to additional modalities (e.g., audio and video) and open-vocabulary entities. We hope that our released code, datasets, and auditing interfaces facilitate reproducible research on scalable, data-centric MMKG enrichment.

\section{Limitations}
\label{sec:Limitations}

Our enrichment quality depends on the reliability of off-the-shelf image-to-text models and fusion LLMs, which may produce incomplete or noisy descriptions for challenging images. We currently generate a single sentence per image and use a single fixed prompt, without systematically studying alternative prompting or multi-aspect captioning strategies. Finally, while we provide an optional human-auditing interface, our main experiments focus on link prediction; broader downstream applications (e.g., QA or entity-centric retrieval) are left for future work.

\begin{credits}
\subsubsection{\ackname} The first author is supported by the China Scholarship Council (NO. 202206540007) and the University of Amsterdam. This funding source had no influence on the study design, data collection, analysis, or manuscript preparation and approval. This work was partially supported by the EU's Horizon Europe programme within the ENEXA project (grant Agreement no. 101070305). This project has received funding from the Horizon Europe research and innovation programme under the Marie Skłodowska-Curie grant agreement No 101146515 KG-PROVENANCE.

\subsubsection{\discintname}
The authors declare that they have no competing interests relevant to the content of this article.
\end{credits}

%
%
%
\bibliographystyle{splncs04}
\bibliography{mybibliography}
%




\end{document}